\documentclass[sigconf, screen, dvipsnames, authorversion]{acmart}

\usepackage{hyperref}
\usepackage[noend]{algorithmic}
\usepackage{algorithm}
\usepackage{multirow}
\usepackage{amsmath}
\usepackage{bbm}
\usepackage{booktabs}
\usepackage[inline]{enumitem} 
\usepackage{subcaption}
\usepackage{bm} 
\usepackage{xcolor}

\usepackage{lipsum}

\usepackage{wrapfig}

\usepackage{arydshln}
\makeatletter
\def\adl@drawiv#1#2#3{
        \hskip.5\tabcolsep
        \xleaders#3{#2.5\@tempdimb #1{1}#2.5\@tempdimb}%
                #2\z@ plus1fil minus1fil\relax
        \hskip.5\tabcolsep}
\newcommand{\cdashlinelr}[1]{%
  \noalign{\vskip\aboverulesep
          \global\let\@dashdrawstore\adl@draw
          \global\let\ adl@draw\adl@drawiv}
  \cdashline{#1}
  \noalign{\global\let\adl@draw\@dashdrawstore
          \vskip\belowrulesep}}
\makeatother

\usepackage{tikz}
\usetikzlibrary{automata, arrows, bayesnet, bending}

\usepackage{tikzscale}

\copyrightyear{2025}
\acmYear{2025}
\setcopyright{rightsretained}
\acmConference[RecSys '25]{Proceedings of the Nineteenth ACM Conference on
Recommender Systems}{September 22--26, 2025}{Prague, Czech Republic}
\acmBooktitle{Proceedings of the Nineteenth ACM Conference on Recommender
Systems (RecSys '25), September 22--26, 2025, Prague, Czech
Republic}\acmDOI{10.1145/3705328.3759308}
\acmISBN{979-8-4007-1364-4/2025/09}

\begin{document}
\title{Meta Off-Policy Estimation}

\author{Olivier Jeunen}
\affiliation{
  \institution{aampe}
  \city{Antwerp}
  \country{Belgium}
}
\email{olivier@aampe.com}

\begin{abstract}
Off-policy estimation (OPE) methods enable unbiased offline evaluation of recommender systems, directly estimating the online reward some target policy would have obtained, from offline data and with statistical guarantees.
The theoretical elegance of the framework combined with practical successes have led to a surge of interest, with many competing estimators now available to practitioners and researchers.
Among these, Doubly Robust methods provide a prominent strategy to combine value- and policy-based estimators.

In this work, we take an alternative perspective to combine a set of OPE estimators and their associated confidence intervals into a single, more accurate estimate.
Our approach leverages a correlated fixed-effects meta-analysis framework, explicitly accounting for dependencies among estimators that arise due to shared data.
This yields a best linear unbiased estimate (\textsc{blue}) of the target policy's value, along with an appropriately conservative confidence interval that reflects inter-estimator correlation.
We validate our method on both simulated and real-world data, demonstrating improved statistical efficiency over existing individual estimators. 
\end{abstract}

\maketitle

\section{Introduction \& Motivation}
Recommender systems power personalisation on the world wide web, in a broad variety of consumer-facing applications and use-cases.
Reliable offline evaluation of such systems has been a prevalent problem, as repeatedly reported in the literature~\cite{Garcin2014,Rossetti2016,Jeunen2018}. 
Recently, recommendations are often seen through a \emph{decision-making} lens~\cite{Joachims2021}.
This enables the use of causal and counterfactual inference methods to derive offline estimators that directly target online success metrics~\cite{Jeunen2021Thesis}.
\emph{Off-policy} estimation methods~\cite{Saito2021,Vasile2020} have seen several practical successes in the recommender systems literature, both for evaluation and learning tasks~\cite{Gilotte2018,Gruson2019,chen2019top,Jeunen2023_C3PO,Jeunen2023_nDCG,Jeunen2024_DeltaOPE,Jeunen2024_MOO}.
As a result, a swath of off-policy estimation methods are available to researchers and practitioners to choose from, with only limited guidance to select an estimator for the task at hand~\cite{Udagawa2023,Felicioni2024}.

Indeed: even if we limit ourselves to (asymptotically) unbiased estimators that leverage the Inverse Propensity Score (IPS), we have the classical IPS estimator~\cite{HorvitzThompson1952}, Self-Normalised IPS (SNIPS)~\cite{Swaminathan2015}, $\beta$-IPS~\cite{Gupta2024}, and Doubly Robust~\cite{Dudik2014}.
These all leverage slightly different signals in the logged data to yield unbiased estimates with Gaussian confidence intervals that exhibit guarantees on statistical coverage. 

The key insight in this work is that these estimators and intervals are complementary.
We can treat them as (correlated) study results, and leverage techniques from statistical meta-analysis~\cite{Cooper2019} to combine them into an estimator that is provably as least as efficient (i.e. has lower variance), whilst retaining unbiasedness.
This is enabled by a classical statistical method to compute the Best Linear Unbiased Estimate (\textsc{blue})~\cite{Aitken1936}, which only requires a vector of means and a covariance matrix for the original input estimators.

\begin{figure}[!t]
    \centering
    \includegraphics[width=\linewidth]{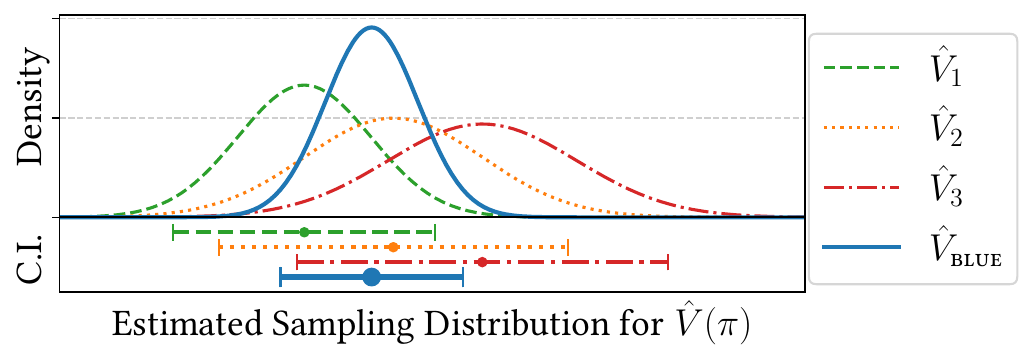}
    \caption{Consider a set of estimators $\{\hat V_1, \hat V_2, \hat V_3\}$ with their corresponding confidence intervals. We combine these estimates to obtain a Best Linear Unbiased Estimate (\textsc{blue}) that retains coverage guarantees whilst reducing variance.}
    \label{fig:intuition}
\end{figure}

Figure~\ref{fig:intuition} visualises the intuition behind our approach: a set of unbiased input estimators $\{\hat V_1, \hat V_2, \hat V_3\}$ is combined to form the unbiased $\hat V_{\textsc{blue}}$, which has the same coverage on a tighter interval.
When biased estimators are used as input, naturally, \textsc{blue} also loses its unbiasedness.
In these cases, \textsc{blue} might still bring performance improvements that stem from the holistic consolidation of complementary individual estimators, trading off bias and variance.

In what follows, we derive the method from first principles and show how it is used in conjunction with common estimators.
We leverage the Delta method to obtain asymptotically unbiased covariance estimates for ratio estimators, such as SNIPS.

We empirically validate the merit of \textsc{blue}, both on a synthetic simulation setup where we change environmental parameters to observe changes in performance, as well as a publicly available dataset for OPE~\cite{Saito2021_OBP}.
All experimental results are fully reproducible, and our source code can be found at \href{https://github.com/olivierjeunen/meta-ope-recsys-2025}{github.com/olivierjeunen/meta-ope-recsys-2025}.

Our approach combines simple, well-established elements from the existing literature to reduce OPE standard errors over the best standalone estimator on the Open Bandit Dataset by over 50\%---equivalent to a fourfold increase in the amount of logged data---whilst incurring minimal additional computational overhead.
Given its effectiveness and simplicity, we expect the \textsc{blue} approach to become part of common practice for robust off-policy evaluation.

\section{Background \& Related Work}
We frame the recommendation task as a decision-making problem, where a context $X$ informs a recommendation policy $\pi$ on which actions $A$ to take: $\pi(a|x)\equiv\mathsf{P}(A=a|X=x;\Pi=\pi)$.
The context typically describes a user, and the action set $\mathcal{A}$ can consist of (sets or rankings~\cite{Gupta2024_WSDM} of) items~\cite{Jeunen2023} or even model weights~\cite{Jeunen2024_MOO}.

Recommendations lead to rewards $R$, which are typically linked to any type of online metric of interest (e.g. clicks, conversions, retention, revenue).
The value of a policy $\pi$ is the expected reward we obtain when exposing it to users: $V(\pi) = \mathbb{E}_{a 
\sim\pi(\cdot|x)}[r]$.

\subsection{Off-Policy Estimation}
Often, we have access to data collected under some logging policy $\pi_0$ (e.g. the current production system), and we want to estimate $V(\pi)$ for some new policy $\pi$ (e.g. a potential update to the system).
Off-policy estimation methods provide tools to estimate this quantity, with statistical guarantees~\cite{Vasile2020,Saito2021}.
Inverse Propensity Score (IPS) weighting~\cite[Ch. 9]{Owen2013} enables unbiased estimation of $V(\pi)$ from a dataset $\mathcal{D}=\{(x_i,a_i,r_i)_{i=1}^{N}\}$ logged under $\pi_0$, as:
\begin{equation}
    \hat V_{\rm IPS}(\pi) = \frac{1}{|\mathcal{D}|} \sum_{(x,a_0,r) \in \mathcal{D}}\frac{\pi(a_0|x)}{\pi_0(a_0|x)}r.
\end{equation}
Whilst unbiased, IPS comes with high variance.
Counter-measures that retain (asymptotic) unbiasedness include the use of multiplicative (i.e. SNIPS~\cite{Swaminathan2015}) or additive (i.e. $\beta$-IPS~\cite{Gupta2024}) control variates.
Alternatively, variance can be traded in for bias by clipping the IPS weights~\cite{Ionides2008, Lichtenberg2023}.
Another approach is the Direct Method (DM), leveraging a reward model $\hat r (a,x)$ to extrapolate to unseen actions~\cite{Jeunen2021_Pessimism}:
\begin{equation}
    \hat V_{\rm DM}(\pi) = \frac{1}{|\mathcal{D}|} \sum_{(x,a_0,r) \in \mathcal{D}} \sum_{a \in \mathcal{A}}\pi(a|x) \hat r (a,x).
\end{equation}
This significantly reduces variance, but almost surely brings bias.
The Doubly Robust (DR) family of approaches combines the strengths of IPS and DM to retain unbiasedness whilst reducing variance~\cite{Dudik2014}:
\begin{gather}
    \hat V_{\rm DR}(\pi) = \nonumber\\
    \frac{1}{|\mathcal{D}|} \sum_{(x,a_0,r) \in \mathcal{D}} \left(\frac{\pi(a_0|x)}{\pi_0(a_0|x)}(r - \hat r (a_0,x)) +   \sum_{a \in \mathcal{A}}\pi(a|x) \hat r (a,x)\right).
\end{gather}
Extensions have been proposed~\cite{Su2019, Su2020,Farajtabar2018} and adopted~\cite{Sagtani2024,Jeunen2023_AuctionGym}, but practical improvements from DR are not guaranteed~\cite{Jeunen2020_DR}.

\subsection{Meta-analysis}
Statistical methods for meta-analysis were first introduced by \citet{Pearson1904}, to collate and aggregate data from independent clinical studies that target the same estimand.
By combining confidence intervals obtained through different studies, a new and more accurate meta-estimate can be obtained, retaining statistical guarantees of confidence interval coverage.
In a ``fixed effect'' model, an assumption is made that all input estimators estimate the exact same underlying population and estimand~\cite{Dekkers2018}.
Whilst this is often an unrealistic assumption when aggregating research results in e.g. medical fields, it aligns well with our intended use-case.
Indeed, our input estimators all target $V(\pi)$.
They will, however, not be independent, as they are typically estimated from the same logged dataset.
\citet{Aitken1936} studied the problem of linearly combining correlated observations, providing the foundation for the methods we build upon.
See \citet[Ch.~19]{Cooper2019} for an in-depth overview.

Recent contemporaneous work proposes \textsc{opera}~\cite{Nie2024}, leveraging an iterative bootstrapping procedure to estimate and constrainedly optimise the mean squared error for an aggregate estimator in general reinforcement learning scenarios---highlighting that it is unclear how to compute covariance among certain estimators.
In contrast, we derive an exact closed-form solution that provably minimises variance without requiring hyperparameters, deriving covariance estimates via the Delta method.
The result is an efficient and effective estimator with distributionally consistent frequentist guarantees on analytically computable confidence intervals.
\section{Methodology \& Contributions}
\subsection{Best Linear Unbiased Estimation}
We aim to linearly combine the results of $K$ unbiased off-policy estimators into a new estimator that retains unbiasedness whilst having minimal variance.
Let $\hat{\bm{\mu}} = (\hat V_1(\pi), \ldots, \hat V_K(\pi))^\intercal$ describe a vector of $K$ estimator means, and $\hat{\bm{\Sigma}} = {\rm Cov}(\hat{\bm{\mu}})$ their $K\times K$ covariance matrix.
It is a well-known result that the Best Linear Unbiased Estimate (\textsc{blue}) is given by~\cite{Aitken1936,Cooper2019}:
\begin{equation}\label{eq:blue}
    \hat{V}_{\textsc{blue}}(\pi) = \frac{\bm{1}^\intercal \hat{\bm{\Sigma}}^{-1} \hat{\bm{\mu}}}{\bm{1}^\intercal \hat{\bm{\Sigma}}^{-1} \bm{1}}, \quad \text{with}\quad \widehat{\rm Var}\left(\hat{V}_{\textsc{blue}}(\pi)\right) = \frac{1}{\bm{1}^\intercal \hat{\bm{\Sigma}}^{-1} \bm{1}}.
\end{equation}
This can be reproduced as the solution of a constrained optimisation problem over weight vectors $\bm w$ with unit sum, to minimise the variance of the resulting estimator.

A Gaussian $(1-\alpha)$\% confidence interval is then given by:
\begin{equation}
\hat{V}_{\textsc{blue}}(\pi) \pm z_{1 - \alpha/2} \cdot \sqrt{\widehat{\text{Var}}\left(\hat{V}_{\textsc{blue}}(\pi)\right)},
\end{equation}
where $z_{1 - \alpha/2}$ is the standard normal critical value.
When all input estimators are unbiased, $\hat{V}_{\textsc{blue}}$ provides the provably optimal (i.e. variance-minimising) linear combination of its inputs (retaining unbiasedness through linearity of expectation).
The resulting variance is upper-bounded by the lowest-variance input estimator.

This simple procedure provides a statistically principled way to combine multiple OPE estimates into a single, interpretable point estimate with a valid uncertainty quantification that reflects inter-estimator dependence through $\Sigma$.
For most common estimators (IPS~\cite{HorvitzThompson1952}, $\beta$-IPS~\cite{Gupta2024}, DR~\cite{Dudik2014}, among others), variances and covariances are estimated through the sample covariance over the logged data $\mathcal{D}$.
When $\hat V(\pi)$ is a \textit{ratio} estimator, this is no longer the case.
Indeed, we need to resort to specialised methods to obtain approximate estimates for these quantities.

\subsection{Covariances for Ratio Estimators}

A common ratio estimator that is popular in OPE applications is the Self-Normalised IPS (SNIPS) estimator~\cite{Swaminathan2015,Gilotte2018,London2023}.
SNIPS leverages a multiplicative control variate, and is defined as a ratio of two sample means.
As such, we can write it as:
\begin{equation}
    \hat V_{\rm SNIPS}(\pi) = \frac{\sum_{(x,a_0,r) \in \mathcal{D}} \frac{\pi(a_0|x)}{\pi_0(a_0|x)} r}{\sum_{(x,a_0,r) \in \mathcal{D} \frac{\pi(a_0|x)}{\pi_0(a_0|x)}}} = \frac{\hat V_{\rm IPS}(\pi)}{\hat V_{\rm SN}(\pi)}.
\end{equation}
Estimates for SNIPS' variance are typically derived through the Delta method~\cite[Ch. 11]{Owen2013}, see e.g.~\cite{Swaminathan2015,Jeunen2024_DeltaOPE}.
In what follows, we apply a similar approximation to additionally estimate the covariance between $\hat V_{\rm SNIPS}$ and any other estimator $\hat V$ when computing $\bm \Sigma$.

The Delta method leverages a first‑order Taylor series expansion around an estimator to approximate the asymptotic sampling distribution of a non-linear transformation of that estimator.
It requires us to compute partial derivatives for $V_{\rm SNIPS}$, as:
\begin{equation}
    \nabla V_{\rm SNIPS} =
    \begin{bmatrix}
        \frac{\partial}{\partial V_{\rm IPS}} \frac{V_{\rm IPS} }{V_{\rm SN}} \\
        \frac{\partial}{\partial V_{\rm SN}} \frac{V_{\rm IPS} }{V_{\rm SN}} 
    \end{bmatrix} = 
    \begin{bmatrix}
         \frac{1 }{V_{\rm SN}} \\
        - \frac{V_{\rm IPS} }{V_{\rm SN^{2}}} 
    \end{bmatrix} .
\end{equation}
This then yields an asymptotically unbiased covariance estimate:
\begin{equation}
    {\rm Cov} \left( \hat V_{\rm SNIPS},\hat V\right) \approx \frac{1}{\hat V_{\rm SN}}   {\rm Cov} \left( \hat V_{\rm IPS},\hat V\right) -  \frac{\hat V_{\rm IPS}}{\hat V_{\rm SN}^2}   {\rm Cov} \left( \hat V_{\rm SN},\hat V\right).
\end{equation}
This can be plugged into $\bm \Sigma$ to be used with Eqs.~\ref{eq:blue}, adding the SNIPS estimator to the set of estimators that \textsc{blue} optimally combines.
\begin{figure*}[!t]
    \centering
    \includegraphics[width=\linewidth]{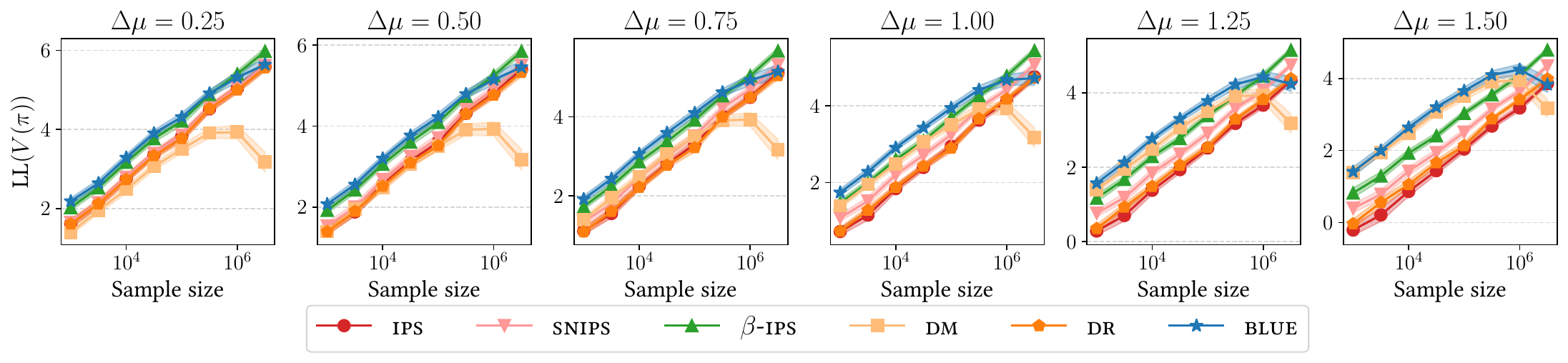}
    \caption{The log-likelihood at the true policy value ${\rm LL}(V(\pi))$, for Gaussian confidence intervals obtained through varying off-policy estimators. We increase the sample size over the $x$-axis and the divergence between the logging and target policies over columns. Unbiased estimators improve as the sample size grows and divergence decreases, the biased  \textsc{dm} estimator outperforms with small samples and large divergences. Our proposed \textsc{blue} approach optimally combines individual estimators in the majority of settings, only suffering in cases where the  \textsc{dm} interval is biased and over-confident.}
    \label{fig:1}
\end{figure*}

\section{Empirical Validation \& Discussion}
The core research question we wish to validate empirically is:
\begin{description}
    \item[\textbf{RQ}] Does \textsc{blue} improve accuracy over individual estimators?
\end{description}
This requires access to a dataset of logged bandit feedback with contextual features, actions, propensities, and rewards.
Furthermore, we require a ground truth policy value to compare against.
Such datasets are scarce~\cite{Lefortier2016,Saito2021_OBP}.
Moreover, they are limited in the insights they can unveil, in that they come with a fixed set of target policies and sample sizes.
Simulations provide a logical alternative in such cases~\cite{Rohde2018}, as they are inherently reproducible and allow us to intervene on environmental parameters to observe the impact on competing methods' performance to gain insights.

As such, we consider both reproducible simulations, as well as real-world data from the Open Bandit Dataset and Pipeline~\cite{Saito2021_OBP}.
All source code to reproduce the results presented in this section is provided at \href{https://github.com/olivierjeunen/meta-ope-recsys-2025}{github.com/olivierjeunen/meta-ope-recsys-2025}.

\subsection{Synthetic Simulation Results}
We instantiate a logging policy $\pi_{0}$, and aim to estimate the value of a target policy $V(\pi)$ using data collected under $\pi_0$.
We vary the sample size that is available to the estimators, as well as the divergence between the logging and target policies $D(\pi_0||\pi)$.

The theoretical expectation is that unbiased (\textsc{ips}-based) estimators improve as the sample size increases, and perform worse as the divergence $D(\pi_0||\pi)$ grows (implying a low effective sample size~\cite{Jeunen2024_MOO,Elvira2022}).
A value-based method like \textsc{dm} will have a different bias-variance trade-off.
For large sample sizes, \textsc{dm} will converge to a low-variance but biased estimate, implying a confidence interval that does not contain the true value $V(\pi)$.
For small sample sizes, the divergence $D(\pi_0||\pi)$ will influence whether \textsc{dm} or \textsc{ips}-based methods are preferable.
When all input estimators are unbiased, $\hat V_{\textsc{blue}}$ will be unbiased too.
When we include \textsc{dm}, $\hat V_{\textsc{blue}}$ loses its unbiasedness but can still exhibit strong performance when \textsc{dm} brings a significant decrease in estimation variance.

\emph{Performance} in this sense implies three desiderata. 
We want an estimator $\hat V(\pi)$ to:
\begin{enumerate*}[label=(\roman*)]
    \item have low error $|\hat V(\pi) - V(\pi)|$ and
    \item low variance ${\rm Var}(\hat V(\pi))$, whilst
    \item retaining coverage for its confidence intervals: $\mathsf{P}\left(V(\pi)\in \left[\hat{V}(\pi) \pm z_{1 - \alpha/2} \cdot \sqrt{\widehat{\text{Var}}\left(\hat{V}(\pi)\right)}\right]\right)=(1-\alpha)$.
\end{enumerate*}

We unify these into a single metric: the log-likelihood of the true target policy value ${\rm LL}(V(\pi))$ under the normal distribution implied by the estimator's confidence interval $\mathcal{N}(\hat V, \widehat{\rm Var}(\hat V))$.
This is the logarithm for the $y$-axis in Figure~\ref{fig:intuition}. Indeed, it is desirable for the sampling distribution of an estimator to tightly concentrate its probability mass near the true value, which ${\rm LL}(V(\pi))$ reflects.

Following recent work---and for ease of implementation and reproducibility---we consider Gaussian policies, which naturally arise when, e.g., modelling scalarisation parameters in multi-objective recommendation settings~\cite{Jeunen2024_MOO,Jeunen2024_DeltaOPE}.
We let the logging policy $\pi_0$ be a standard normal $\mathcal{N}(0,1)$, and vary the target policy $\pi\equiv\mathcal{N}(\Delta\mu,1)$ to move further away from $\pi_0$.
For simplicity but without loss of generality, we define the reward as $\mathsf{P}(R|A=a)=\mathcal{N}(a,1)$.
The reward model $\hat r$ used by \textsc{dm} and \textsc{dr} is a biased and noisy estimator of this reward: $\hat r(a) \sim \mathcal{N}(a+0.0025,2)$.
Note that the injection of noise into $\hat r$ is a common approach to represent approximate uncertainty~\cite{Gal2016}, and necessary in our setting to obtain confidence intervals for \textsc{dm} that do not collapse instantly.

Figure~\ref{fig:1} visualises 95\% confidence intervals around ${\rm LL}(V(\pi))$ for all estimators as the sample size increases over the $x$-axis, and the divergence $\Delta\mu$ increases over the columns, across $2^8$ different random seeds.
Empirical observations align with our theoretical expectations: $\beta$-\textsc{ips} uses the variance-minimising constant additive control variate successfully---with \textsc{dm} outperforming when $D(\pi_0||\pi)$ is high, but converging to a biased estimate which harms performance at large sample sizes.
The meta-analytic \textsc{blue} estimator manages to aggregate complementary information from individual estimators (\textsc{snips}, $\beta$-\textsc{ips}, \textsc{dm}, \textsc{dr}), and outperform them in the majority of cases.
Nevertheless, a bias-variance trade-off is apparent.
When \textsc{dm}'s bias is the driving factor in its error rather than variance---occurring at the inflection point in the plots for a sample size of roughly one million---its overconfidence harms \textsc{blue} as well.
This is to be expected, as the variance of the \textsc{blue} estimate is upper-bounded by that of its lowest input: \textsc{dm}.
This implies that, as \textsc{blue}'s ${\rm LL}(V(\pi))$ is higher, the estimator's error is improved significantly.
Both in settings with lower $D(\pi_0||\pi)$ (when $\beta$-\textsc{ips} is optimal) and those with higher $D(\pi_0||\pi)$ (when \textsc{dm} is optimal), \textsc{blue} successfully identifies the optimal component among its inputs, yielding a consistently optimal estimator.

In this simple non-contextual simulated setting, the heterogeneity of information that is encoded in different estimators is limited.
As a result, the performance gains that we can expect from \textsc{blue} are constrained as well.
When we do not include \textsc{dm}, \textsc{blue} recovers $\beta$-\textsc{ips} with a marginal improvement that is not practically significant.
This changes for richer datasets and use-cases---which we can additionally use to provide ablation study results for \textsc{blue}.

\begin{figure}[!t]
    \centering
    \includegraphics[width=\linewidth]{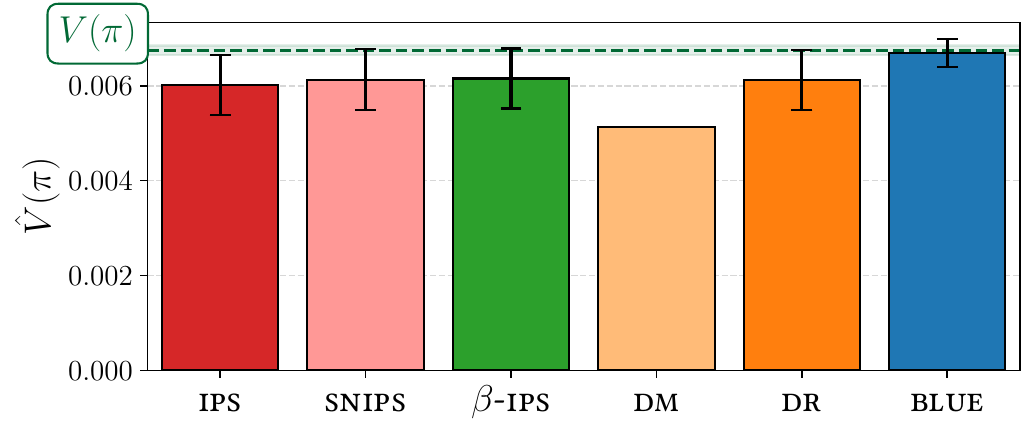}
    \caption{Estimation results on the Men's campaign from the Open Bandit Dataset~\cite{Saito2021_OBP}, visualising 99\% confidence intervals for various estimators as well as the true value $V(\pi)$. Our proposed \textsc{blue} approach significantly improves estimation accuracy over existing individual estimators.}
    \label{fig:2}
\end{figure}

\subsection{Open Bandit Dataset and Pipeline}
The Open Bandit Pipeline provides a Python package for off-policy evaluation, bringing ease of implementation and reproducibility~\cite{Saito2021}.
It includes a real-world logged bandit dataset from ZOZOTOWN, where top-3 lists of recommendations were shown to users, and all information to recover $x$, $a$, $r$ and $\pi_0(a|x)$ is provided.
It includes data collected under a random logging policy and a Bernoulli Thompson sampling target policy $\pi$~\cite{Chapelle2011}, where propensities are estimated via Monte Carlo sampling~\cite{Jeunen2025_TS}.
Through the data collected under $\pi$, we can obtain a Monte Carlo estimate for $V(\pi)$ along with its variance.
Using any of the aforementioned off-policy estimators and data from $\pi_0$, we can obtain an interval for $\hat V(\pi)$.
Figure~\ref{fig:2} visualises these intervals, with \textsc{dm} and \textsc{dr} leveraging a random forest classifier to estimate rewards~\cite{Breiman2001,Pedregosa2011}.
Empirical observations again corroborate theory: \textsc{blue} remains unbiased, but with significantly reduced variance, leading to a tighter confidence interval around the true value.
The width of \textsc{blue}'s confidence interval is down to 47\% of that of the lowest-variance input estimator (\textsc{dr}).
Note that the estimator mean for \textsc{blue} is not a simple interpolation from its inputs, and that it successfully leverages the covariance structure $\bm \Sigma$ to increase the \textsc{blue} estimate and bring it closer to $V(\pi)$.
We additionally note that results were qualitatively similar for the other campaigns in the ZOZOTOWN dataset---albeit with less pronounced improvements due to already well-performing base estimators.

\begin{figure}[!t]
    \centering
    \includegraphics[width=\linewidth]{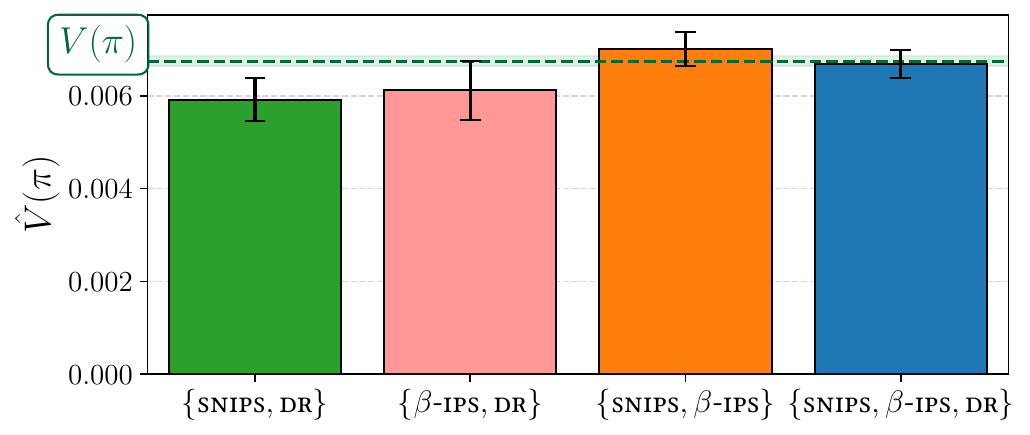}
    \caption{Ablation study results when withholding estimator information from \textsc{blue}. These show that all of \textsc{snips}, $\beta$-\textsc{ips} and \textsc{dr} contribute to \textsc{blue}'s final performance.}
    \label{fig:2_ablation}
\end{figure}

\textit{Ablation study results.}
A natural follow-up question to consider, is which of \textsc{blue}'s input estimators exhibit the largest effect on the performance of the resulting combined estimator.
As such, we follow the same setup to compute the \textsc{blue} on subsets of available estimators.
Figure~\ref{fig:2_ablation} visualises these ablation study results.
Since \textsc{ips} is a special case of $\beta$-\textsc{ips} with $\beta\equiv 0$, these estimators will be highly correlated and potentially lead to an ill-conditioned covariance matrix $\hat{\bm\Sigma}$.
Direct use of the reward model via \textsc{dm} lacks uncertainty quantification, leading to an apparent and problematic bias.
As such, we include three (asymptotically) unbiased but complementary estimators: $\beta$-\textsc{ips}, \textsc{snips} and \textsc{dr}.

We observe that the removal of $\beta$-\textsc{ips} has a negative impact on performance.
The combination of \textsc{snips} and \textsc{dr} remains valid, but both \textsc{snips} itself and our covariance estimates are only \textit{asymptotically} unbiased.
As a result, the finite-sample performance of the combined estimator may exhibit considerable variability.

The removal of the \textsc{dr} estimator is least impactful.
Nevertheless, the best linear unbiased combination of all three estimators provides the tightest confidence interval as well as the lowest estimation error measured as the distance between the true policy value and the estimator mean.
These empirical insights highlight the merit of our proposed approach.

\section{Conclusions \& Outlook}
Off-policy estimation methods are widely used by researchers and practitioners to---among other use-cases---perform an unbiased offline evaluation of their recommender system.
Several competing estimators exist, which can complicate the task at hand.
Our work leverages the insight that multiple unbiased estimators can entail complementary information, and that this information can be combined to form a new estimator with appropriately conservative confidence intervals.
To achieve this, the covariance among existing estimators must be quantified, to then inform a best linear unbiased estimate for the target policy's value.
We provide simple and efficient methods for doing so, and empirically validate that our approach significantly improves the statistical efficiency of standalone estimators---both on reproducible simulations as well as publicly available real-world recommendation data.

Our results demonstrate that a simple application of existing ideas from the meta-analysis literature to OPE problems can yield substantial improvements to estimation precision.
In our experiments, \textsc{blue} provides equivalent benefits to the availability of a $4\times$ increase in the size of the logged data, whilst requiring minimal additional computation.
These results imply a significant practical impact for the use of OPE methods in real-world scenarios.

Our \textsc{blue} approach relies on a single matrix inversion of $K$ dimensions, followed by several matrix-vector products.
As these are all efficiently computable and differentiable, a natural avenue for future work is to apply it to general off-policy learning objectives, where estimator variance remains a well-known challenge.

We believe that this opens up promising avenues for future research, applying off-policy estimators for real-world successes.

\bibliographystyle{ACM-Reference-Format}
\bibliography{bibliography}


\begin{thebibliography}{47}


\ifx \showCODEN    \undefined \def \showCODEN     #1{\unskip}     \fi
\ifx \showISBNx    \undefined \def \showISBNx     #1{\unskip}     \fi
\ifx \showISBNxiii \undefined \def \showISBNxiii  #1{\unskip}     \fi
\ifx \showISSN     \undefined \def \showISSN      #1{\unskip}     \fi
\ifx \showLCCN     \undefined \def \showLCCN      #1{\unskip}     \fi
\ifx \shownote     \undefined \def \shownote      #1{#1}          \fi
\ifx \showarticletitle \undefined \def \showarticletitle #1{#1}   \fi
\ifx \showURL      \undefined \def \showURL       {\relax}        \fi
\providecommand\bibfield[2]{#2}
\providecommand\bibinfo[2]{#2}
\providecommand\natexlab[1]{#1}
\providecommand\showeprint[2][]{arXiv:#2}

\bibitem[Aitken(1936)]%
        {Aitken1936}
\bibfield{author}{\bibinfo{person}{Alexander~C. Aitken}.} \bibinfo{year}{1936}\natexlab{}.
\newblock \showarticletitle{IV.—On Least Squares and Linear Combination of Observations}.
\newblock \bibinfo{journal}{\emph{Proc. of the Royal Society of Edinburgh}}  \bibinfo{volume}{55} (\bibinfo{year}{1936}), \bibinfo{pages}{42–48}.
\newblock


\bibitem[Breiman(2001)]%
        {Breiman2001}
\bibfield{author}{\bibinfo{person}{Leo Breiman}.} \bibinfo{year}{2001}\natexlab{}.
\newblock \showarticletitle{Random Forests}.
\newblock \bibinfo{journal}{\emph{Machine Learning}} \bibinfo{volume}{45}, \bibinfo{number}{1} (\bibinfo{year}{2001}), \bibinfo{pages}{5--32}.
\newblock


\bibitem[Chapelle and Li(2011)]%
        {Chapelle2011}
\bibfield{author}{\bibinfo{person}{Olivier Chapelle} {and} \bibinfo{person}{Lihong Li}.} \bibinfo{year}{2011}\natexlab{}.
\newblock \showarticletitle{An Empirical Evaluation of Thompson Sampling}. In \bibinfo{booktitle}{\emph{Advances in Neural Information Processing Systems}}, Vol.~\bibinfo{volume}{24}. \bibinfo{publisher}{Curran Associates, Inc.}
\newblock


\bibitem[Chen et~al\mbox{.}(2019)]%
        {chen2019top}
\bibfield{author}{\bibinfo{person}{Minmin Chen}, \bibinfo{person}{Alex Beutel}, \bibinfo{person}{Paul Covington}, \bibinfo{person}{Sagar Jain}, \bibinfo{person}{Francois Belletti}, {and} \bibinfo{person}{Ed~H Chi}.} \bibinfo{year}{2019}\natexlab{}.
\newblock \showarticletitle{Top-k off-policy correction for a REINFORCE recommender system}. In \bibinfo{booktitle}{\emph{Proc. of the Twelfth ACM International Conference on Web Search and Data Mining}}. \bibinfo{pages}{456--464}.
\newblock


\bibitem[Cooper et~al\mbox{.}(2019)]%
        {Cooper2019}
\bibfield{author}{\bibinfo{person}{Harris Cooper}, \bibinfo{person}{Larry~V Hedges}, {and} \bibinfo{person}{Jeffrey~C Valentine}.} \bibinfo{year}{2019}\natexlab{}.
\newblock \bibinfo{booktitle}{\emph{The handbook of research synthesis and meta-analysis}}.
\newblock \bibinfo{publisher}{Russell Sage Foundation}.
\newblock


\bibitem[Dekkers(2018)]%
        {Dekkers2018}
\bibfield{author}{\bibinfo{person}{Olaf~M Dekkers}.} \bibinfo{year}{2018}\natexlab{}.
\newblock \showarticletitle{Meta-analysis: Key features, potentials and misunderstandings}.
\newblock \bibinfo{journal}{\emph{Res Pract Thromb Haemost}} \bibinfo{volume}{2}, \bibinfo{number}{4} (\bibinfo{date}{Oct.} \bibinfo{year}{2018}), \bibinfo{pages}{658--663}.
\newblock


\bibitem[Dudík et~al\mbox{.}(2014)]%
        {Dudik2014}
\bibfield{author}{\bibinfo{person}{Miroslav Dudík}, \bibinfo{person}{Dumitru Erhan}, \bibinfo{person}{John Langford}, {and} \bibinfo{person}{Lihong Li}.} \bibinfo{year}{2014}\natexlab{}.
\newblock \showarticletitle{Doubly Robust Policy Evaluation and Optimization}.
\newblock \bibinfo{journal}{\emph{Statist. Sci.}} \bibinfo{volume}{29}, \bibinfo{number}{4} (\bibinfo{year}{2014}), \bibinfo{pages}{485--511}.
\newblock
\showISSN{08834237, 21688745}


\bibitem[Elvira et~al\mbox{.}(2022)]%
        {Elvira2022}
\bibfield{author}{\bibinfo{person}{Víctor Elvira}, \bibinfo{person}{Luca Martino}, {and} \bibinfo{person}{Christian~P. Robert}.} \bibinfo{year}{2022}\natexlab{}.
\newblock \showarticletitle{Rethinking the Effective Sample Size}.
\newblock \bibinfo{journal}{\emph{International Statistical Review}} \bibinfo{volume}{90}, \bibinfo{number}{3} (\bibinfo{year}{2022}), \bibinfo{pages}{525--550}.
\newblock


\bibitem[Farajtabar et~al\mbox{.}(2018)]%
        {Farajtabar2018}
\bibfield{author}{\bibinfo{person}{Mehrdad Farajtabar}, \bibinfo{person}{Yinlam Chow}, {and} \bibinfo{person}{Mohammad Ghavamzadeh}.} \bibinfo{year}{2018}\natexlab{}.
\newblock \showarticletitle{More Robust Doubly Robust Off-policy Evaluation}. In \bibinfo{booktitle}{\emph{Proc. of the 35th International Conference on Machine Learning}} \emph{(\bibinfo{series}{Proc. of Machine Learning Research}, Vol.~\bibinfo{volume}{80})}. \bibinfo{publisher}{PMLR}, \bibinfo{pages}{1447--1456}.
\newblock
\urldef\tempurl%
\url{https://proceedings.mlr.press/v80/farajtabar18a.html}
\showURL{%
\tempurl}


\bibitem[Felicioni et~al\mbox{.}(2024)]%
        {Felicioni2024}
\bibfield{author}{\bibinfo{person}{Nicolò Felicioni}, \bibinfo{person}{Michael Benigni}, {and} \bibinfo{person}{Maurizio~Ferrari Dacrema}.} \bibinfo{year}{2024}\natexlab{}.
\newblock \bibinfo{title}{Automated Off-Policy Estimator Selection via Supervised Learning}.
\newblock
\showeprint[arxiv]{2406.18022}~[cs.LG]


\bibitem[Gal and Ghahramani(2016)]%
        {Gal2016}
\bibfield{author}{\bibinfo{person}{Yarin Gal} {and} \bibinfo{person}{Zoubin Ghahramani}.} \bibinfo{year}{2016}\natexlab{}.
\newblock \showarticletitle{Dropout as a Bayesian Approximation: Representing Model Uncertainty in Deep Learning}. In \bibinfo{booktitle}{\emph{Proc. of The 33rd International Conference on Machine Learning}} \emph{(\bibinfo{series}{Proc. of Machine Learning Research}, Vol.~\bibinfo{volume}{48})}. \bibinfo{publisher}{PMLR}, \bibinfo{pages}{1050--1059}.
\newblock


\bibitem[Garcin et~al\mbox{.}(2014)]%
        {Garcin2014}
\bibfield{author}{\bibinfo{person}{Florent Garcin}, \bibinfo{person}{Boi Faltings}, \bibinfo{person}{Olivier Donatsch}, \bibinfo{person}{Ayar Alazzawi}, \bibinfo{person}{Christophe Bruttin}, {and} \bibinfo{person}{Amr Huber}.} \bibinfo{year}{2014}\natexlab{}.
\newblock \showarticletitle{Offline and Online Evaluation of News Recommender Systems at Swissinfo.Ch}. In \bibinfo{booktitle}{\emph{Proc. of the 8th ACM Conference on Recommender Systems}} \emph{(\bibinfo{series}{RecSys '14})}. \bibinfo{publisher}{ACM}, \bibinfo{pages}{169–176}.
\newblock
\showISBNx{9781450326681}
\urldef\tempurl%
\url{https://doi.org/10.1145/2645710.2645745}
\showURL{%
\tempurl}


\bibitem[Gilotte et~al\mbox{.}(2018)]%
        {Gilotte2018}
\bibfield{author}{\bibinfo{person}{Alexandre Gilotte}, \bibinfo{person}{Cl\'{e}ment Calauz\`{e}nes}, \bibinfo{person}{Thomas Nedelec}, \bibinfo{person}{Alexandre Abraham}, {and} \bibinfo{person}{Simon Doll\'{e}}.} \bibinfo{year}{2018}\natexlab{}.
\newblock \showarticletitle{Offline {A/B} Testing for Recommender Systems}. In \bibinfo{booktitle}{\emph{Proc. of the Eleventh ACM International Conference on Web Search and Data Mining}} \emph{(\bibinfo{series}{WSDM '18})}. \bibinfo{publisher}{ACM}, \bibinfo{pages}{198–206}.
\newblock
\showISBNx{9781450355810}
\urldef\tempurl%
\url{https://doi.org/10.1145/3159652.3159687}
\showURL{%
\tempurl}


\bibitem[Gruson et~al\mbox{.}(2019)]%
        {Gruson2019}
\bibfield{author}{\bibinfo{person}{Alois Gruson}, \bibinfo{person}{Praveen Chandar}, \bibinfo{person}{Christophe Charbuillet}, \bibinfo{person}{James McInerney}, \bibinfo{person}{Samantha Hansen}, \bibinfo{person}{Damien Tardieu}, {and} \bibinfo{person}{Ben Carterette}.} \bibinfo{year}{2019}\natexlab{}.
\newblock \showarticletitle{Offline Evaluation to Make Decisions About Playlist Recommendation Algorithms}. In \bibinfo{booktitle}{\emph{Proc. of the Twelfth ACM International Conference on Web Search and Data Mining}} \emph{(\bibinfo{series}{WSDM '19})}. \bibinfo{publisher}{ACM}, \bibinfo{pages}{420–428}.
\newblock
\showISBNx{9781450359405}
\href{https://doi.org/10.1145/3289600.3291027}{doi:\nolinkurl{10.1145/3289600.3291027}}


\bibitem[Gupta et~al\mbox{.}(2024a)]%
        {Gupta2024_WSDM}
\bibfield{author}{\bibinfo{person}{Shashank Gupta}, \bibinfo{person}{Philipp Hager}, \bibinfo{person}{Jin Huang}, \bibinfo{person}{Ali Vardasbi}, {and} \bibinfo{person}{Harrie Oosterhuis}.} \bibinfo{year}{2024}\natexlab{a}.
\newblock \showarticletitle{Unbiased Learning to Rank: On Recent Advances and Practical Applications}. In \bibinfo{booktitle}{\emph{Proc. of the 17th ACM International Conference on Web Search and Data Mining}} \emph{(\bibinfo{series}{WSDM '24})}. \bibinfo{publisher}{ACM}, \bibinfo{pages}{1118–1121}.
\newblock
\showISBNx{9798400703713}
\href{https://doi.org/10.1145/3616855.3636451}{doi:\nolinkurl{10.1145/3616855.3636451}}


\bibitem[Gupta et~al\mbox{.}(2024b)]%
        {Gupta2024}
\bibfield{author}{\bibinfo{person}{Shashank Gupta}, \bibinfo{person}{Olivier Jeunen}, \bibinfo{person}{Harrie Oosterhuis}, {and} \bibinfo{person}{Maarten de Rijke}.} \bibinfo{year}{2024}\natexlab{b}.
\newblock \showarticletitle{Optimal Baseline Corrections for Off-Policy Contextual Bandits}. In \bibinfo{booktitle}{\emph{Proc. of the 18th ACM Conference on Recommender Systems}} \emph{(\bibinfo{series}{RecSys '24})}. \bibinfo{publisher}{ACM}, \bibinfo{pages}{722–732}.
\newblock
\showISBNx{9798400705052}


\bibitem[Horvitz and Thompson(1952)]%
        {HorvitzThompson1952}
\bibfield{author}{\bibinfo{person}{Daniel~G. Horvitz} {and} \bibinfo{person}{Donovan~J. Thompson}.} \bibinfo{year}{1952}\natexlab{}.
\newblock \showarticletitle{A Generalization of Sampling Without Replacement From a Finite Universe}.
\newblock \bibinfo{journal}{\emph{J. Amer. Statist. Assoc.}} \bibinfo{volume}{47}, \bibinfo{number}{260} (\bibinfo{year}{1952}), \bibinfo{pages}{663--685}.
\newblock
\showISSN{01621459, 1537274X}
\urldef\tempurl%
\url{http://www.jstor.org/stable/2280784}
\showURL{%
\tempurl}


\bibitem[Ionides(2008)]%
        {Ionides2008}
\bibfield{author}{\bibinfo{person}{Edward~L. Ionides}.} \bibinfo{year}{2008}\natexlab{}.
\newblock \showarticletitle{Truncated Importance Sampling}.
\newblock \bibinfo{journal}{\emph{Journal of Computational and Graphical Statistics}} \bibinfo{volume}{17}, \bibinfo{number}{2} (\bibinfo{year}{2008}), \bibinfo{pages}{295--311}.
\newblock


\bibitem[Jeunen(2021)]%
        {Jeunen2021Thesis}
\bibfield{author}{\bibinfo{person}{Olivier Jeunen}.} \bibinfo{year}{2021}\natexlab{}.
\newblock \emph{\bibinfo{title}{Offline Approaches to Recommendation with Online Success}}.
\newblock \bibinfo{thesistype}{Ph.\,D. Dissertation}. \bibinfo{school}{University of Antwerp}.
\newblock


\bibitem[Jeunen(2023)]%
        {Jeunen2023_C3PO}
\bibfield{author}{\bibinfo{person}{Olivier Jeunen}.} \bibinfo{year}{2023}\natexlab{}.
\newblock \showarticletitle{A Probabilistic Position Bias Model for Short-Video Recommendation Feeds}. In \bibinfo{booktitle}{\emph{Proc. of the 17th ACM Conference on Recommender Systems}} \emph{(\bibinfo{series}{RecSys '23})}. \bibinfo{publisher}{ACM}, \bibinfo{pages}{675–681}.
\newblock
\showISBNx{9798400702419}
\href{https://doi.org/10.1145/3604915.3608777}{doi:\nolinkurl{10.1145/3604915.3608777}}


\bibitem[Jeunen(2025)]%
        {Jeunen2025_TS}
\bibfield{author}{\bibinfo{person}{Olivier Jeunen}.} \bibinfo{year}{2025}\natexlab{}.
\newblock \bibinfo{title}{Counterfactual Inference under Thompson Sampling}.
\newblock
\showeprint[arxiv]{2504.08773}~[cs.IR]


\bibitem[Jeunen and Goethals(2020)]%
        {Jeunen2020_DR}
\bibfield{author}{\bibinfo{person}{Olivier Jeunen} {and} \bibinfo{person}{Bart Goethals}.} \bibinfo{year}{2020}\natexlab{}.
\newblock \showarticletitle{An Empirical Evaluation of Doubly Robust Learning for Recommendation}. In \bibinfo{booktitle}{\emph{REVEAL Workshop at ACM RecSys '20}} \emph{(\bibinfo{series}{REVEAL '20})}.
\newblock


\bibitem[Jeunen and Goethals(2021)]%
        {Jeunen2021_Pessimism}
\bibfield{author}{\bibinfo{person}{Olivier Jeunen} {and} \bibinfo{person}{Bart Goethals}.} \bibinfo{year}{2021}\natexlab{}.
\newblock \showarticletitle{Pessimistic Reward Models for Off-Policy Learning in Recommendation}. In \bibinfo{booktitle}{\emph{Proc. of the 15th ACM Conference on Recommender Systems}} \emph{(\bibinfo{series}{RecSys '21})}. \bibinfo{publisher}{ACM}, \bibinfo{pages}{63–74}.
\newblock
\showISBNx{9781450384582}
\href{https://doi.org/10.1145/3460231.3474247}{doi:\nolinkurl{10.1145/3460231.3474247}}


\bibitem[Jeunen and Goethals(2023)]%
        {Jeunen2023}
\bibfield{author}{\bibinfo{person}{Olivier Jeunen} {and} \bibinfo{person}{Bart Goethals}.} \bibinfo{year}{2023}\natexlab{}.
\newblock \showarticletitle{Pessimistic Decision-Making for Recommender Systems}.
\newblock \bibinfo{journal}{\emph{ACM Trans. Recomm. Syst.}} \bibinfo{volume}{1}, \bibinfo{number}{1}, Article \bibinfo{articleno}{4} (\bibinfo{date}{feb} \bibinfo{year}{2023}), \bibinfo{numpages}{27}~pages.
\newblock


\bibitem[Jeunen et~al\mbox{.}(2024a)]%
        {Jeunen2024_MOO}
\bibfield{author}{\bibinfo{person}{Olivier Jeunen}, \bibinfo{person}{Jatin Mandav}, \bibinfo{person}{Ivan Potapov}, \bibinfo{person}{Nakul Agarwal}, \bibinfo{person}{Sourabh Vaid}, \bibinfo{person}{Wenzhe Shi}, {and} \bibinfo{person}{Aleksei Ustimenko}.} \bibinfo{year}{2024}\natexlab{a}.
\newblock \showarticletitle{Multi-Objective Recommendation via Multivariate Policy Learning}. In \bibinfo{booktitle}{\emph{Proc. of the 18th ACM Conference on Recommender Systems}} \emph{(\bibinfo{series}{RecSys '24})}. \bibinfo{publisher}{ACM}, \bibinfo{pages}{712–721}.
\newblock
\showISBNx{9798400705052}
\href{https://doi.org/10.1145/3640457.3688132}{doi:\nolinkurl{10.1145/3640457.3688132}}


\bibitem[Jeunen et~al\mbox{.}(2023)]%
        {Jeunen2023_AuctionGym}
\bibfield{author}{\bibinfo{person}{Olivier Jeunen}, \bibinfo{person}{Sean Murphy}, {and} \bibinfo{person}{Ben Allison}.} \bibinfo{year}{2023}\natexlab{}.
\newblock \showarticletitle{Off-Policy Learning-to-Bid with AuctionGym}. In \bibinfo{booktitle}{\emph{Proc. of the 29th ACM SIGKDD Conference on Knowledge Discovery and Data Mining}} \emph{(\bibinfo{series}{KDD '23})}. \bibinfo{publisher}{ACM}, \bibinfo{pages}{4219–4228}.
\newblock
\showISBNx{9798400701030}


\bibitem[Jeunen et~al\mbox{.}(2024b)]%
        {Jeunen2023_nDCG}
\bibfield{author}{\bibinfo{person}{Olivier Jeunen}, \bibinfo{person}{Ivan Potapov}, {and} \bibinfo{person}{Aleksei Ustimenko}.} \bibinfo{year}{2024}\natexlab{b}.
\newblock \showarticletitle{On (Normalised) Discounted Cumulative Gain as an Off-Policy Evaluation Metric for Top-n Recommendation}. In \bibinfo{booktitle}{\emph{Proc. of the 30th ACM SIGKDD Conference on Knowledge Discovery and Data Mining}} \emph{(\bibinfo{series}{KDD '24})}. \bibinfo{publisher}{ACM}, \bibinfo{pages}{1222–1233}.
\newblock
\showISBNx{9798400704901}
\href{https://doi.org/10.1145/3637528.3671687}{doi:\nolinkurl{10.1145/3637528.3671687}}


\bibitem[Jeunen and Ustimenko(2024)]%
        {Jeunen2024_DeltaOPE}
\bibfield{author}{\bibinfo{person}{Olivier Jeunen} {and} \bibinfo{person}{Aleksei Ustimenko}.} \bibinfo{year}{2024}\natexlab{}.
\newblock \showarticletitle{$\Delta$-OPE: Off-Policy Estimation with Pairs of Policies}. In \bibinfo{booktitle}{\emph{Proc. of the 18th ACM Conference on Recommender Systems}} \emph{(\bibinfo{series}{RecSys '24})}. \bibinfo{publisher}{ACM}, \bibinfo{pages}{878–883}.
\newblock
\showISBNx{9798400705052}
\href{https://doi.org/10.1145/3640457.3688162}{doi:\nolinkurl{10.1145/3640457.3688162}}


\bibitem[Jeunen et~al\mbox{.}(2018)]%
        {Jeunen2018}
\bibfield{author}{\bibinfo{person}{Olivier Jeunen}, \bibinfo{person}{Koen Verstrepen}, {and} \bibinfo{person}{Bart Goethals}.} \bibinfo{year}{2018}\natexlab{}.
\newblock \showarticletitle{Fair Offline Evaluation Methodologies for Implicit-Feedback Recommender Systems with MNAR Data}. In \bibinfo{booktitle}{\emph{Workshop on Offline Evaluation for Recommender Systems at RecSys '18}} \emph{(\bibinfo{series}{REVEAL '18})}.
\newblock


\bibitem[Joachims et~al\mbox{.}(2021)]%
        {Joachims2021}
\bibfield{author}{\bibinfo{person}{Thorsten Joachims}, \bibinfo{person}{Ben London}, \bibinfo{person}{Yi Su}, \bibinfo{person}{Adith Swaminathan}, {and} \bibinfo{person}{Lequn Wang}.} \bibinfo{year}{2021}\natexlab{}.
\newblock \showarticletitle{Recommendations as Treatments}.
\newblock \bibinfo{journal}{\emph{AI Magazine}} \bibinfo{volume}{42}, \bibinfo{number}{3} (\bibinfo{date}{Nov.} \bibinfo{year}{2021}), \bibinfo{pages}{19--30}.
\newblock


\bibitem[Lefortier et~al\mbox{.}(2016)]%
        {Lefortier2016}
\bibfield{author}{\bibinfo{person}{Damien Lefortier}, \bibinfo{person}{Adith Swaminathan}, \bibinfo{person}{Xiaotao Gu}, \bibinfo{person}{Thorsten Joachims}, {and} \bibinfo{person}{Maarten de Rijke}.} \bibinfo{year}{2016}\natexlab{}.
\newblock \showarticletitle{Large-scale Validation of Counterfactual Learning Methods: A Test-Bed}. In \bibinfo{booktitle}{\emph{NIPS What If Workshop on Inference and Learning of Hypothetical and Counterfactual Interventions in Complex Systems}}.
\newblock
\showeprint[arxiv]{1612.00367}~[cs.LG]


\bibitem[Lichtenberg et~al\mbox{.}(2023)]%
        {Lichtenberg2023}
\bibfield{author}{\bibinfo{person}{Jan~Malte Lichtenberg}, \bibinfo{person}{Alexander Buchholz}, \bibinfo{person}{Giuseppe~Di Benedetto}, \bibinfo{person}{Matteo Ruffini}, {and} \bibinfo{person}{Ben London}.} \bibinfo{year}{2023}\natexlab{}.
\newblock \showarticletitle{Double Clipping: Less-Biased Variance Reduction in Off-Policy Evaluation}. In \bibinfo{booktitle}{\emph{CONSEQUENCES Workshop at ACM RecSys '23}} \emph{(\bibinfo{series}{CONSEQUENCES '23})}.
\newblock
\showeprint[arxiv]{2309.01120}~[cs.LG]


\bibitem[London et~al\mbox{.}(2023)]%
        {London2023}
\bibfield{author}{\bibinfo{person}{Ben London}, \bibinfo{person}{Alexander Buchholz}, \bibinfo{person}{Giuseppe Di~Benedetto}, \bibinfo{person}{Jan~Malte Lichtenberg}, \bibinfo{person}{Yannik Stein}, {and} \bibinfo{person}{Thorsten Joachims}.} \bibinfo{year}{2023}\natexlab{}.
\newblock \showarticletitle{Self-Normalized Off-Policy Estimators for Ranking}. In \bibinfo{booktitle}{\emph{CONSEQUENCES Workshop at ACM RecSys '23}} \emph{(\bibinfo{series}{CONSEQUENCES '23})}.
\newblock


\bibitem[Nie et~al\mbox{.}(2024)]%
        {Nie2024}
\bibfield{author}{\bibinfo{person}{Allen Nie}, \bibinfo{person}{Yash Chandak}, \bibinfo{person}{Christina~J. Yuan}, \bibinfo{person}{Anirudhan Badrinath}, \bibinfo{person}{Yannis Flet-Berliac}, {and} \bibinfo{person}{Emma Brunskill}.} \bibinfo{year}{2024}\natexlab{}.
\newblock \showarticletitle{OPERA: Automatic Offline Policy Evaluation with Re-weighted Aggregates of Multiple Estimators}. In \bibinfo{booktitle}{\emph{Advances in Neural Information Processing Systems}}, \bibfield{editor}{\bibinfo{person}{A.~Globerson}, \bibinfo{person}{L.~Mackey}, \bibinfo{person}{D.~Belgrave}, \bibinfo{person}{A.~Fan}, \bibinfo{person}{U.~Paquet}, \bibinfo{person}{J.~Tomczak}, {and} \bibinfo{person}{C.~Zhang}} (Eds.), Vol.~\bibinfo{volume}{37}. \bibinfo{publisher}{Curran Associates, Inc.}, \bibinfo{pages}{103652--103680}.
\newblock


\bibitem[Owen(2013)]%
        {Owen2013}
\bibfield{author}{\bibinfo{person}{Art~B. Owen}.} \bibinfo{year}{2013}\natexlab{}.
\newblock \bibinfo{booktitle}{\emph{Monte Carlo theory, methods and examples}}.
\newblock


\bibitem[Pearson(1904)]%
        {Pearson1904}
\bibfield{author}{\bibinfo{person}{Karl Pearson}.} \bibinfo{year}{1904}\natexlab{}.
\newblock \showarticletitle{Report on Certain Enteric Fever Inoculation Statistics}.
\newblock \bibinfo{journal}{\emph{BMJ}} \bibinfo{volume}{2}, \bibinfo{number}{2288} (\bibinfo{year}{1904}), \bibinfo{pages}{1243--1246}.
\newblock
\showISSN{0007-1447}
\href{https://doi.org/10.1136/bmj.2.2288.1243}{doi:\nolinkurl{10.1136/bmj.2.2288.1243}}


\bibitem[Pedregosa et~al\mbox{.}(2011)]%
        {Pedregosa2011}
\bibfield{author}{\bibinfo{person}{Fabian Pedregosa}, \bibinfo{person}{Ga{{\"e}}l Varoquaux}, \bibinfo{person}{Alexandre Gramfort}, \bibinfo{person}{Vincent Michel}, \bibinfo{person}{Bertrand Thirion}, \bibinfo{person}{Olivier Grisel}, \bibinfo{person}{Mathieu Blondel}, \bibinfo{person}{Peter Prettenhofer}, \bibinfo{person}{Ron Weiss}, \bibinfo{person}{Vincent Dubourg}, \bibinfo{person}{Jake Vanderplas}, \bibinfo{person}{Alexandre Passos}, \bibinfo{person}{David Cournapeau}, \bibinfo{person}{Matthieu Brucher}, \bibinfo{person}{Matthieu Perrot}, {and} \bibinfo{person}{{{\'E}}douard Duchesnay}.} \bibinfo{year}{2011}\natexlab{}.
\newblock \showarticletitle{Scikit-learn: Machine Learning in Python}.
\newblock \bibinfo{journal}{\emph{Journal of Machine Learning Research}} \bibinfo{volume}{12}, \bibinfo{number}{85} (\bibinfo{year}{2011}), \bibinfo{pages}{2825--2830}.
\newblock
\urldef\tempurl%
\url{http://jmlr.org/papers/v12/pedregosa11a.html}
\showURL{%
\tempurl}


\bibitem[Rohde et~al\mbox{.}(2018)]%
        {Rohde2018}
\bibfield{author}{\bibinfo{person}{David Rohde}, \bibinfo{person}{Stephen Bonner}, \bibinfo{person}{Travis Dunlop}, \bibinfo{person}{Flavian Vasile}, {and} \bibinfo{person}{Alexandros Karatzoglou}.} \bibinfo{year}{2018}\natexlab{}.
\newblock \showarticletitle{RecoGym: A Reinforcement Learning Environment for the problem of Product Recommendation in Online Advertising}. In \bibinfo{booktitle}{\emph{RecSys REVEAL Workshop on Offline Evaluation for Recommender Systems}}.
\newblock


\bibitem[Rossetti et~al\mbox{.}(2016)]%
        {Rossetti2016}
\bibfield{author}{\bibinfo{person}{Marco Rossetti}, \bibinfo{person}{Fabio Stella}, {and} \bibinfo{person}{Markus Zanker}.} \bibinfo{year}{2016}\natexlab{}.
\newblock \showarticletitle{Contrasting Offline and Online Results When Evaluating Recommendation Algorithms}. In \bibinfo{booktitle}{\emph{Proc. of the 10th ACM Conference on Recommender Systems}} \emph{(\bibinfo{series}{RecSys '16})}. \bibinfo{publisher}{ACM}, \bibinfo{pages}{31–34}.
\newblock
\showISBNx{9781450340359}


\bibitem[Sagtani et~al\mbox{.}(2024)]%
        {Sagtani2024}
\bibfield{author}{\bibinfo{person}{Hitesh Sagtani}, \bibinfo{person}{Madan~Gopal Jhawar}, \bibinfo{person}{Rishabh Mehrotra}, {and} \bibinfo{person}{Olivier Jeunen}.} \bibinfo{year}{2024}\natexlab{}.
\newblock \showarticletitle{Ad-load Balancing via Off-policy Learning in a Content Marketplace}. In \bibinfo{booktitle}{\emph{Proc. of the 17th ACM International Conference on Web Search and Data Mining}} \emph{(\bibinfo{series}{WSDM '24})}. \bibinfo{publisher}{ACM}, \bibinfo{pages}{586–595}.
\newblock
\showISBNx{9798400703713}
\href{https://doi.org/10.1145/3616855.3635846}{doi:\nolinkurl{10.1145/3616855.3635846}}


\bibitem[Saito et~al\mbox{.}(2021)]%
        {Saito2021_OBP}
\bibfield{author}{\bibinfo{person}{Yuta Saito}, \bibinfo{person}{Shunsuke Aihara}, \bibinfo{person}{Megumi Matsutani}, {and} \bibinfo{person}{Yusuke Narita}.} \bibinfo{year}{2021}\natexlab{}.
\newblock \showarticletitle{Open Bandit Dataset and Pipeline: Towards Realistic and Reproducible Off-Policy Evaluation}. In \bibinfo{booktitle}{\emph{Proc. of the Neural Information Processing Systems Track on Datasets and Benchmarks}}, Vol.~\bibinfo{volume}{1}.
\newblock


\bibitem[Saito and Joachims(2021)]%
        {Saito2021}
\bibfield{author}{\bibinfo{person}{Yuta Saito} {and} \bibinfo{person}{Thorsten Joachims}.} \bibinfo{year}{2021}\natexlab{}.
\newblock \showarticletitle{Counterfactual Learning and Evaluation for Recommender Systems: Foundations, Implementations, and Recent Advances}. In \bibinfo{booktitle}{\emph{Proc. of the 15th ACM Conference on Recommender Systems}} \emph{(\bibinfo{series}{RecSys '21})}. \bibinfo{publisher}{ACM}, \bibinfo{pages}{828–830}.
\newblock
\showISBNx{9781450384582}
\href{https://doi.org/10.1145/3460231.3473320}{doi:\nolinkurl{10.1145/3460231.3473320}}


\bibitem[Su et~al\mbox{.}(2020)]%
        {Su2020}
\bibfield{author}{\bibinfo{person}{Yi Su}, \bibinfo{person}{Maria Dimakopoulou}, \bibinfo{person}{Akshay Krishnamurthy}, {and} \bibinfo{person}{Miroslav Dudik}.} \bibinfo{year}{2020}\natexlab{}.
\newblock \showarticletitle{Doubly robust off-policy evaluation with shrinkage}. In \bibinfo{booktitle}{\emph{Proc. of the 37th International Conference on Machine Learning}} \emph{(\bibinfo{series}{Proc. of Machine Learning Research}, Vol.~\bibinfo{volume}{119})}. \bibinfo{publisher}{PMLR}, \bibinfo{pages}{9167--9176}.
\newblock
\urldef\tempurl%
\url{https://proceedings.mlr.press/v119/su20a.html}
\showURL{%
\tempurl}


\bibitem[Su et~al\mbox{.}(2019)]%
        {Su2019}
\bibfield{author}{\bibinfo{person}{Yi Su}, \bibinfo{person}{Lequn Wang}, \bibinfo{person}{Michele Santacatterina}, {and} \bibinfo{person}{Thorsten Joachims}.} \bibinfo{year}{2019}\natexlab{}.
\newblock \showarticletitle{{CAB}: Continuous Adaptive Blending for Policy Evaluation and Learning}. In \bibinfo{booktitle}{\emph{Proc. of the 36th International Conference on Machine Learning}} \emph{(\bibinfo{series}{Proc. of Machine Learning Research}, Vol.~\bibinfo{volume}{97})}. \bibinfo{publisher}{PMLR}, \bibinfo{pages}{6005--6014}.
\newblock


\bibitem[Swaminathan and Joachims(2015)]%
        {Swaminathan2015}
\bibfield{author}{\bibinfo{person}{Adith Swaminathan} {and} \bibinfo{person}{Thorsten Joachims}.} \bibinfo{year}{2015}\natexlab{}.
\newblock \showarticletitle{The Self-Normalized Estimator for Counterfactual Learning}. In \bibinfo{booktitle}{\emph{Advances in Neural Information Processing Systems}}, Vol.~\bibinfo{volume}{28}. \bibinfo{publisher}{Curran Associates, Inc.}
\newblock


\bibitem[Udagawa et~al\mbox{.}(2023)]%
        {Udagawa2023}
\bibfield{author}{\bibinfo{person}{Takuma Udagawa}, \bibinfo{person}{Haruka Kiyohara}, \bibinfo{person}{Yusuke Narita}, \bibinfo{person}{Yuta Saito}, {and} \bibinfo{person}{Kei Tateno}.} \bibinfo{year}{2023}\natexlab{}.
\newblock \showarticletitle{Policy-Adaptive Estimator Selection for Off-Policy Evaluation}.
\newblock \bibinfo{journal}{\emph{Proc. of the AAAI Conference on Artificial Intelligence}} \bibinfo{volume}{37}, \bibinfo{number}{8} (\bibinfo{date}{Jun.} \bibinfo{year}{2023}), \bibinfo{pages}{10025--10033}.
\newblock


\bibitem[Vasile et~al\mbox{.}(2020)]%
        {Vasile2020}
\bibfield{author}{\bibinfo{person}{Flavian Vasile}, \bibinfo{person}{David Rohde}, \bibinfo{person}{Olivier Jeunen}, {and} \bibinfo{person}{Amine Benhalloum}.} \bibinfo{year}{2020}\natexlab{}.
\newblock \showarticletitle{A Gentle Introduction to Recommendation as Counterfactual Policy Learning}. In \bibinfo{booktitle}{\emph{Proc. of the 28th ACM Conference on User Modeling, Adaptation and Personalization}} \emph{(\bibinfo{series}{UMAP '20})}. \bibinfo{publisher}{ACM}, \bibinfo{pages}{392–393}.
\newblock
\showISBNx{9781450368612}
\href{https://doi.org/10.1145/3340631.3398666}{doi:\nolinkurl{10.1145/3340631.3398666}}


\end{thebibliography}

\end{document}